# USE OF CONVOLUTIONAL NEURAL NETWORKS IN DETECTION OF BRANCH DIEBACK DISEASE IN PEQUI TREES

Ronilson R. Pinho [a,b], Eluã R. Coutinho [d], Nilza de L. P. Sales [c], Sildenir A. Ribeiro [b], Sergio Manuel Serra da Cruz [a]

[a] *Federal University of Rio de Janeiro – UFRJ- Institute of Computing – IC - Postgraduate Program in Informatics - PPGI- Rio de Janeiro, 20010-974, Brazil*

[b] *Federal Center for Technological Education Celso Suckow da Fonseca- CEFET-RJ-BSI- Rio de Janeiro, RJ, 20785-220, Brazil*

[c] *Federal University of Lavras – UFLA - Postgraduate Program in Forestry Engineering – PPGEF- Lavras, MG, 37200-900, Brazil*

[d] *Fluminense Federal University – UFF - Instituto de Ciência e Tecnologia – Rio das Ostras - RJ - 28895-532, Brazil*

*E-mail: ronilson.pinho@cefet-rj.br; sergioserra@ic.ufrj.br; sildenir.ribeiro@cefet-rj.br; nilzasales@ufla.br; eluacoutinho@id.uff.br*

**Abstract**

This study evaluates and compares five convolutional neural network (CNN) architectures, AlexNet, Inception, LeNet, VGG16, and MobileNet, for diagnosing branch dieback ("*seca-dos-ponteiros"*) in pequi trees (*Caryocar brasiliense*). This phytopathology represents a severe threat to tree development and fruit yield, causing substantial economic and ecological losses across the Brazilian Cerrado. The methodology implements computer vision techniques using deep learning frameworks to process and evaluate a field dataset expanded to 6,272 multi-scale images, captured via UAV and handheld sensors under uncontrolled illumination. The models were trained to detect pathological visual patterns across leaves, branches, and canopies. The experimental initial results reveal that architectural complexity did not guarantee diagnostic reliability, heavily parameterized networks experienced optimization failures (VGG16 collapsed into majority-class prediction), whereas compact architectures achieved greater stability. MobileNet delivered the most competitive overall performance, obtaining the highest Precision (0.71) and ROC-AUC (0.63), closely followed by LeNet (AUC = 0.62). These findings indicate the viability of lightweight CNNs as foundational tools for automated phytosanitary surveillance of native flora in resource-constrained field environments.

## 1. Introduction

Branch dieback poses a significant threat to Brazilian pequi trees (*Caryocar brasiliense*), severely compromising their vegetative development and productivity. This disease drastically reduces the productivity of the plant capacity by impairing fruit set and quality, leading to substantial yield losses in both fruit and timber while undermining the economic and ecological viability of the crop. Over time, symptoms can progress to severe wilting and tissue necrosis, ultimately resulting in tree mortality. Given the severity of this condition, timely identification, targeted management, and conservation strategies are essential to preserve this key botanical resource of the Brazilian Cerrado biome (Doerl, 2021).

In recent years, deep learning approaches have been increasingly deployed in smart farming to address challenges such as pest detection, yield estimation, and foliar disease diagnosis (Hema et al., 2021). However, most existing literature focuses predominantly on commercial monocultures (such as soybean, fruits, corn, and citrus), leaving wild native species largely overlooked. Thus, investigating Brazilian native flora fills a significant gap in this domain.

Conventional machine learning pipelines often struggle with real-time deployment and feature extraction under complex field conditions. In contrast, deep learning methods offer robust feature representation, providing an effective framework for automated agricultural decision-support systems.

However, visual diagnosis of canopy diseases remains challenging, as early foliar and woody symptoms are often subtle or visually occluded (Sena Junior et al., 2003; Barbedo et al., 2014). In pequi trees affected by dieback, diagnosis is further complicated because affected trees tend to shed their upper leaves prematurely, leaving behind bare, desiccated terminal branches.

To address this challenge, computer vision models based on Convolutional Neural Networks (CNNs) offer a promising automated solution. By training on annotated imagery of leaves, branches, and flower spikes, CNNs can capture diagnostic visual patterns, such as localized discolorations and textural shifts (Javed and Ali, 2020). Specifically, evaluating five pre-trained architectures such as AlexNet, LeNet, VGG16, Inception, and MobileNet enables rapid, high-accuracy discrimination between healthy and infected tissues, supporting timely agronomic interventions.

### *1.1 Related work*

Recent studies have actively explored CNNs for automated plant disease diagnosis. For instance, Jakjoud et al. (2019) deployed the VGGNet architecture for leaf image classification, achieving 90% accuracy while optimizing inference speed on a resource-constrained Raspberry Pi 3 Model B. Similarly, Kerre and Muchiri (2022) evaluated GoogLeNet, ResNet, and VGG to diagnose strawberry leaf spot, attaining accuracies between 90% and 98% and demonstrating the feasibility of deep transfer learning for foliar diseases. Evaluating broader benchmarking, Singh et al. (2019) compared architectures including VGG16, InceptionV3, ResNet, and MobileNet across datasets collected in both controlled and uncontrolled environments; however, they highlighted that deep backbones incur substantial computational overhead, limiting deployment on low-cost edge hardware.

In contrast to prior work focused predominantly on complex, compute-intensive backbones, our study systematically evaluates a balanced spectrum of architectures, ranging from lightweight and classical models (LeNet, MobileNet, and AlexNet) to deeper representations (VGG16 and Inception), to assess trade-offs for edge and low-resource agricultural scenarios. Furthermore, we address class imbalance by implementing targeted data augmentation pipelines, enhancing model generalization and diagnostic robustness under variable field conditions.

## 2. General Concepts

### *2.1 Geographic Distribution and Socioeconomic Relevance*

The pequi tree (*Caryocar brasiliense*) is a typical species of the Brazilian Cerrado, legally protected to foster environmental conservation and safeguard its natural ecosystem (Prance and Silva, 1973). Widely distributed across the Central-West region and federal states including Minas Gerais, Tocantins, Rondônia, São Paulo, and Ceará, the species holds substantial ecological and cultural importance. Its fruits are central to regional cuisine, underpinning local extractivist economies and cultural heritage.

Environmental legislation aims to balance biome conservation with sustainable community development. Morphological features of the pequi tree, including typical leaves and canopy structures, are shown in Figures 1(a), (b), and (c).

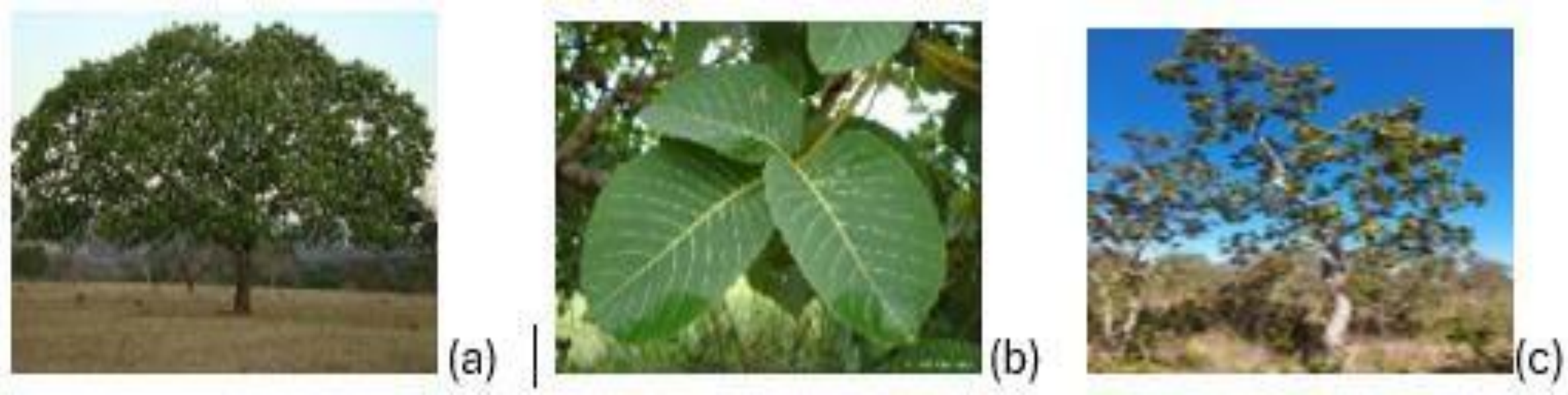

Fig. 1. Images of pequi trees (a and c) and pequi healthy leaves (b).

*2.2 Branch Dieback Syndrome*

Branch dieback in pequi trees is a complex disease driven by the interplay of abiotic and biotic stressors. Contributing factors range from adverse microclimatic conditions and poor edaphic fertility to secondary infestations by insect pests, fungal complexes, and bacterial pathogens. The disorder can affect trees at any developmental stage and is characterized by progressive defoliation coupled with basipetal necrosis that proceeds from the terminal twigs toward older wood.

As symptom severity increases, necrosis advances downward from the terminal twigs into structural scaffolding branches and the main trunk. Without intervention, systemic decline occurs, resulting in severe canopy dieback and eventual tree mortality. Representative progression stages are illustrated in Figure 2: (a) moderate symptom severity, (b) severe canopy defoliation, and (c) extensive structural necrosis leading to tree mortality.

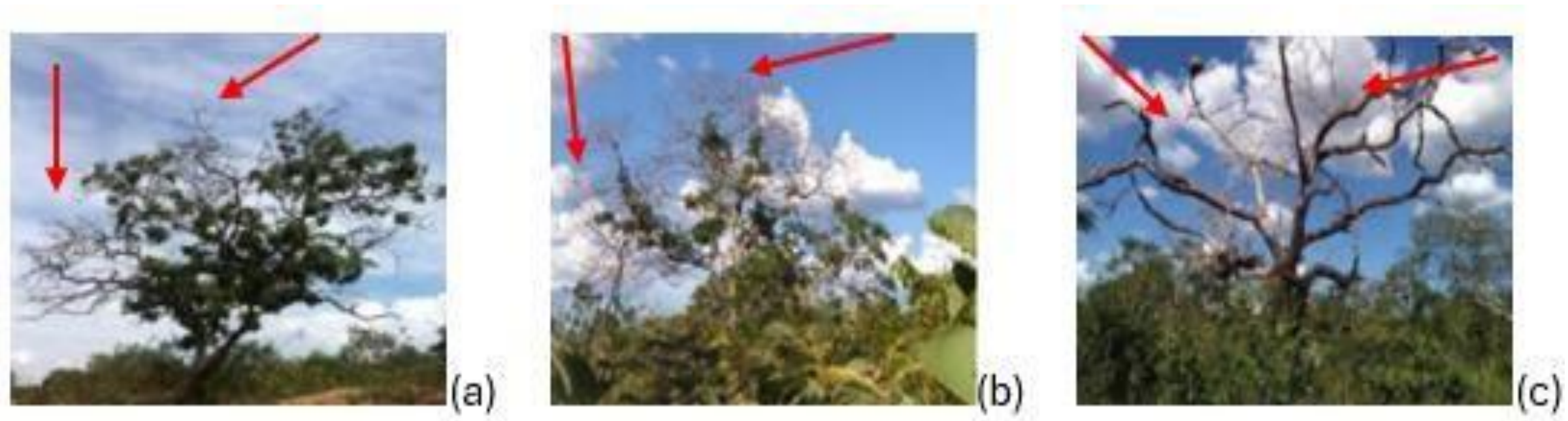

Fig.2. Different levels of branch dieback in pequi trees. Source: (Doerl, 2021).

*2.3 Convolutional Neural Networks*

CNNs represent a specialized class of deep neural networks designed primarily for visual recognition tasks, including image classification, object detection, and regression. Inspired by the biological visual cortex, CNNs rely on a hierarchical representation learning scheme, enabling them to progressively extract features ranging from low-level edges and textures to high-level semantic patterns.

A core component of these networks is the discrete convolution operation, wherein learnable spatial filters (kernels) slide across the input tensor to produce local feature maps. As the network trains via backpropagation, these filter weights adjust to capture domain-specific diagnostic patterns. A standard CNN architecture comprises three primary functional layers:

- Convolutional layers: Apply sets of learnable kernels over the input imagery, incorporating nonlinear activation functions (such as ReLU) to generate activation or feature maps corresponding to distinct visual representations.
- Pooling layers: Perform spatial downsampling on feature maps to reduce dimensional complexity, lower computational latency, and mitigate overfitting. Operations such as max pooling (capturing localized dominant activations) and average pooling (computing mean regional responses) provide translation and shift invariance while retaining prominent features.

- Dense (Fully Connected) layers: Flatten the spatial feature representations into a one-dimensional vector, integrating extracted spatial signatures to generate the final classification logits or regression outputs. In modern architectures, these may be coupled with or replaced by global average pooling to minimize parameter count.

The overall pipeline and representative layer interactions are illustrated in Figure 3.

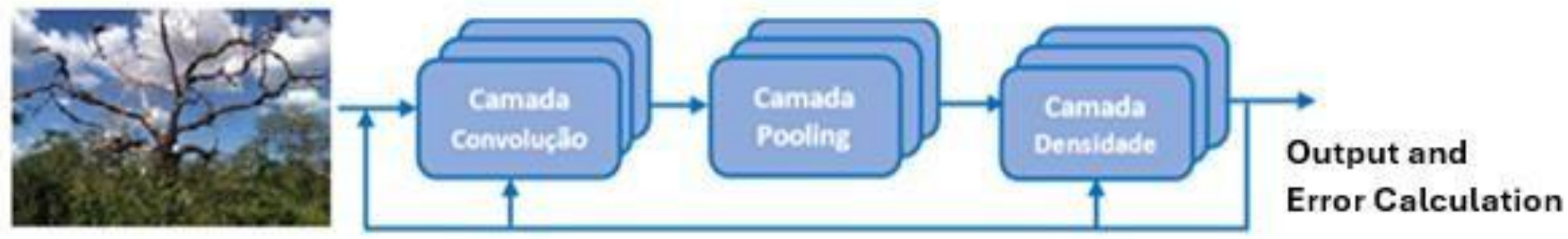

Fig. 3. Processing a CNN - Adaptation (Hema et al, 2021)

*2.3.1 Visual Geometry Group Network - VGG16*

VGG16 is a foundational deep convolutional neural network developed by Simonyan and Zisserman (2014) as part of the Visual Geometry Group (VGGNet) family. Originally evaluated on the ImageNet Large Scale Visual Recognition Challenge (ILSVRC), the architecture is characterized by its architectural simplicity, stacking homogeneous $3 \times 3$ convolution kernels with stride 1 throughout all feature-extraction stages, interleaved with $2 \times 2$ max-pooling layers with stride 2 for spatial downsampling. In this work, an adapted VGG16 topology was implemented to process input images downscaled to $64 \times 64 \times 3$, structured across five hierarchical convolutional blocks and a custom classification head:

- Blocks 1 and 2 - Consist of two consecutive $3 \times 3$ convolutional layers each (yielding 64 and 128 feature channels, respectively), followed by $2 \times 2$ max pooling to downsample spatial dimensions to $32 \times 32$ and $16 \times 16$.

- Blocks 3, 4, and 5 - Employ three consecutive $3 \times 3$ convolutions each, expanding the channel depth to 256, 512, and 512 feature maps while downsampling spatial resolutions to $8 \times 8$, $4 \times 4$, and $2 \times 2$, respectively.

- Classification Head - The final $2 \times 2 \times 512$ feature maps are flattened into a 2,048-dimensional vector, followed by two fully connected layers of 4,096 units regularized via dropout ($p = 0.5$). An intermediate dense layer of 1,000 units connects to the final 2-unit softmax output layer configured for binary classification (healthy vs. diseased).

The complete layer-by-layer architectural summary, output shapes, and parameter counts are detailed in Table 1.

**Table 1**. VGG16 with its 16 layers.

| Layer (type) | Output shape | Param |
|---|---|---|
| conv2d (Conv2D) | (None,64,64,64) | 640 |
| conv2d_1 (Conv2D) | (None, 64,64,64) | 36,928 |
| max_pooling2d (MaxPooling2D) | (None, 32,32,64) | 0 |
| conv2d_2 (Conv2D) | (None,32,32,128) | 73,856 |
| conv2d_3 (Conv2D) | (None,32,32,128) | 147,584 |
| max_pooling2d_1(MaxPooling2D) | (None, 16,16,128) | 0 |
| conv2d_4 (Conv2D) | (None, 16,16,256) | 295,168 |
| conv2d_5 (Conv2D) | (None,16,16,256) | 590,080 |
| conv2d_6 (Conv2D) | (None,16,16,256) | 590,080 |
| max_pooling2d_2 (MaxPooling2D) | (None,8,8,256) | 0 |
| conv2d_7 ( Conv2D) | (None,8,8,512) | 1,180,160 |
| conv2d_8 (Conv2D) | (None,8,8,512) | 2,359,808 |
| conv2d_9 (Conv2D) | (None,8,8,512) | 2,359,808 |
| max_pooling2d_3 (MaxPooling2D) | (None,4,4,512) | 0 |
| Conv2d_10 (Conv2D) | (None,4,4,512) | 2,359,808 |
| Conv2d_11(Conv2D) | (None,4,4,512) | 2,359,808 |
| Conv2d_12 (Conv2D) | (None,4,4,512) | 2,359,808 |
| Max_pooling2d_4 (MaxPooling2D) | (None,2,2,512) | 0 |
| Flatten (Flatten) | (None,2048) | 0 |
| Dense (Dense) | (None,4096) | 8,392,704 |
| Dropout | (None,4096) | 0 |
| Dense_1 (Dense) | (None,4096) | 16,781,312 |
| Dropout_1 (Dropout) | (None,4096) | 0 |
| Dense_2 (Dense) | (None,1000) | 4,097,000 |
| Dense_3(Dense) | (None,2) | 2,002 |

*2.3.2 AlexNet*

AlexNet is a pioneering convolutional neural network architecture introduced by Krizhevsky et al. (2012) that established deep learning as the dominant paradigm in computer vision. The canonical architecture comprises eight learned layers: five convolutional stages (interleaved with max pooling and local response normalization) and three fully connected layers. A key innovation of AlexNet was the adoption of the Rectified Linear Unit (ReLU) activation function, which significantly accelerated gradient convergence compared to saturating nonlinearities such as sigmoid or hyperbolic tangent. In this study, an adapted AlexNet pipeline was evaluated to process input images, structured into sequential feature-extraction and classification stages:

- Convolutional Backbone - The first stage (conv2d_13, 96 filters of size 11×11) maps inputs to 55×55 feature representations, followed by non-overlapping max pooling (max_pooling2d_5, downsampling to 27×27). The second stage (conv2d_14, 256 filters) is downsampled to 13×13 via max_pooling2d_6. The remaining stages comprise three consecutive convolutional layers (conv2d_15, conv2d_16, and conv2d_17 with 384, 384, and 256 filters, respectively), followed by a final pooling operation (max_pooling2d_7) yielding a 6×6×256 representation tensor.

- Fully Connected Classifier - The spatial tensor is vectorized into a 9,216-dimensional vector via flatten_1. The classification head consists of two dense layers of 4,096 units each (dense_4 and

dense_5) regularized with dropout, followed by an intermediate 1,000-unit projection layer (dense_6) and a final 2-unit softmax output layer (dense_7) for binary disease classification.

The complete layer sequence, output tensor shapes, and parameter distributions are summarized in Table 2.

**Table 2**. AlexNet with its layers, generated by the application

| Layer (type) | Output shape | Param |
|---|---|---|
| Conv2d_13 (Conv2D) | (None,55,55,96) | 11,712 |
| Max_pooling2d_5(MaxPooling2D) | (None,27,27,96) | 0 |
| Conv2d_14 (Conv2D) | (None,27,27,256) | 614,656 |
| Conv2d_15(Conv2D) | (None,13,13,384) | 885,120 |
| Conv2d_16(Convd2D) | (None,13,13,384) | 1,327,488 |
| Conv2d_17(Convd2D) | (None,13,13,256) | 894,992 |
| Max_pooling2d_7(MaxPooling2D) | (None,6,6,256) | 0 |
| Flatten_1 (Flatten) | (None,9216) | 0 |
| Dense_4 (Dense) | (None,4096) | 37,752,832 |
| Dense_5( Dense) | (None,4096) | 16,781,312 |
| Dense_6 (Dense) | (None,1000) | 4,097,000 |
| Dense_7 (Dense) | (None,2) | 2,002 |

*2.3.3 Incepton*

The Inception family of CNN, originally introduced by Google via GoogLeNet/Inception-v1 (Szegedy et al., 2015), was designed to improve computational efficiency while scaling network depth and width. A hallmark of the architecture is the Inception block, which processes visual features across multiple spatial resolutions concurrently using parallel branches of $1 \times 1$, $3 \times 3$, and $5 \times 5$ convolutions alongside pooling operations. To mitigate computational bottlenecks, $1 \times 1$ convolutions serve as dimensionality reduction layers before applying larger spatial kernels. Subsequent iterations (Inception-v2 through Inception-v4) introduced factorized asymmetric convolutions and incorporated residual connections (Inception-ResNet) to stabilize gradient propagation in deeper topologies. In this work, an adapted Inception topology was implemented to process single-channel grayscale inputs ($150 \times 150 \times 1$). The architecture comprises three primary functional stages:

- Preliminary Stem - A series of five initial convolutional layers (conv2d through conv2d_4) with intermediate max-pooling operations (max_pooling2d, max_pooling2d_1) extracts low-level textural representations and downsamples spatial dimensions from $150 \times 150$ to $16 \times 16$, expanding depth to 192 feature channels.

- Parallel Inception Blocks - Two successive multi-branch Inception modules extract multi-scale spatial representations at a constant resolution of $16 \times 16$. The first block concatenates parallel branch projections into 256 channels (concatenate), while the second module outputs 288 concatenated channels (concatenate_1).

- Classification Head - The final $16 \times 16 \times 288$ tensor is flattened into a 73,728-dimensional vector (flatten). This representation feeds into a dense layer of 1,024 units (Dense, containing 75,498,496 trainable parameters) regularized by dropout stages (dropout, dropout_1), ending in a 2-unit softmax classifier (dense_1) for binary disease diagnosis.

The comprehensive layer configuration, intermediate tensor dimensions, and parameter allocations are summarized in Table 3.

**Table 3.** Inception with its layers, generated by the application

| Layer (type) | Output shape | Param |
|---|---|---|
| input_layer (Inputlayer) | (None, 150,150, 1) | 0 |
| Conv2d (Conv2d) | (None, 74,74,32) | 320 |
| Conv2d_1 (Conv2d) | (None, 72,72,32) | 9,248 |
| Conv2d_2 (Conv2d) | (None, 72,72,64) | 18,496 |
| max_pooling2D( maxpooling2d) | (None,35,35,64) | 0 |
| Conv2d_3 (Conv2d) | (None,35,35,80) | 5,200 |
| Conv2d_4 (Conv2d) | (None,33,33,192) | 138,432 |
| Max_pooling2d_1 (maxpooling2d) | (None,16,16,192) | 0 |
| Conv2d_6 (Conv2d) | (None,16,16,48) | 9,264 |
| Conv2d_8 (Conv2d) | (None,16,16,64) | 12,254 |
| max_pooling2d_2 (maxpooling2d) | (None,16,16,192) | 0 |
| Conv2d_5 (Conv2d) | (None,16,16,64) | 12,352 |
| Conv2d_7 (Conv2d) | (None,16,16,64) | 27,712 |
| Conv2d_9 (Conv2d) | (None,16,16,96) | 153,696 |
| Conv2d_10 (Conv2d) | (None,16,16,32) | 6,176 |
| Concatenate | (None, 16,16,256) | 0 |
| Conv2d_12 (Conv2d) | (None, 16,16,48) | 12,336 |
| Conv2d_14 (Conv2d) | (None, 16,16,64) | 16,448 |
| max_pooling2d_3 | (None,16,16,256) | 0 |
| Conv2d_11 (Conv2d) | (None,16,16,64) | 16,448 |
| Conv2d_13 (Conv2d) | (None,16,16,64) | 27,712 |
| Conv2d_15 (Conv2d) | (None,16,16,96) | 153,696 |
| Conv2d_16 (Conv2d) | (None,16,16,64) | 16,448 |
| Concatenate_1 (Concatenate) | (None,16,16,288) | 0 |
| Flatten (Flatten) | (None, 73728) | 0 |
| Dropout | (None,73728) | 0 |
| Dense (Dense) | (None, 1024) | 75,498,496 |
| Dropout_1 (Dropout) | (None,1024) | 0 |
| Dense_1 (Dense) | (None, 2) | 2,050 |

*2.3.4 LeNet*

LeNet-5 is a foundational convolutional neural network introduced by LeCun et al. (1998), initially designed for handwritten digit recognition on the MNIST benchmark. The architecture follows a classical, lightweight topology consisting of alternating convolutional and spatial downsampling layers, terminated by fully connected stages. In this study, an adapted LeNet-5 model was evaluated using single-channel grayscale inputs resized to $32 \times 32 \times 1$. As summarized in Table 4, the sequential pipeline is structured into feature-extraction and classification blocks:

- Feature Extraction - The initial layer (conv2d_17, six $5 \times 5$ filters) maps inputs into six $28 \times 28$ feature maps (156 parameters), downsampled to $14 \times 14$ by the first max-pooling stage (max_pooling2d_4). The second convolutional stage (conv2d_18, sixteen $5 \times 5$ filters) expands channel depth to sixteen $10 \times 10$ feature maps (2,416 parameters), which are subsequently reduced to $5 \times 5$ by max_pooling2d_5.

- Classification Head - The resulting spatial tensor ( $5 \times 5 \times 16$ ) is flattened into a 400-dimensional vector (flatten_1). This vector feeds into two sequential fully connected layers with 120 and 84 hidden units (dense_2 with 48,120 parameters; dense_3 with 10,164 parameters), culminating in a 2-unit softmax output layer (dense_4, 170 parameters) for binary disease classification.

**Table 4** - LeNet with its layers, generated by the application

| Layer (type) | Output shape | Param |
|---|---|---|
| Conv2d_17 (Conv2D) | (None,28,28,6) | 156 |
| max_pooling2d_4(MaxPooling2D) | (None,14,14,6) | 0 |
| Conv2d_18 (Conv2D) | (None,10,10,16) | 2,416 |
| max_pooling2d_5(MaxPooling2D) | (None,5,5,16) | 0 |
| Flatten_1 (Flatten) | (None,400) | 0 |
| Dense_2 (Dense) | (None,120) | 48,120 |
| Dense_3 (Dense) | (None,84) | 10,164 |
| Dense_4 (Dense) | (None,2) | 170 |

*2.3.5 Lightweight Mobile-Oriented Architecture*

MobileNet architectures, originally proposed by Howard et al. (2017), are designed to optimize inference latency and memory footprints in embedded and resource-constrained environments. By prioritizing compact topologies and parameter efficiency, such models enable real-time on-device diagnosis without relying on continuous cloud connectivity. In this study, a compact, mobile-oriented convolutional pipeline was implemented to evaluate lightweight classification under edge constraints (Table 5). The sequential architecture processes downsampled inputs through four feature-extraction blocks followed by a dense classification head:

- Convolutional Feature Extractor - Comprises four sequential stages of standard $3 \times 3$ convolutions interleaved with $2 \times 2$ max-pooling layers (max_pooling2d through max_pooling2d_3). Representation depth progressively scales from 32 to 128 feature channels while spatial dimensions are downsampled from $74 \times 74$ to $4 \times 4$.

- Classification Head - The final $4 \times 4 \times 128$ spatial tensor is vectorized via Flatten into a 2,048-dimensional feature representation. This vector feeds into a fully connected layer of 512 neurons (dense, 1,049,088 parameters), culminating in a 2-unit softmax output layer (dense_1, 1,026 parameters) for binary disease classification.

**Table 5** - MobileNet, with its layers

| Layer (type) | Output shape | Param |
|---|---|---|
| Conv2d (Conv2D) | (None,74,74,32) | 320 |
| max_pooling2d (MaxPooling2D) | (None,37,37,32) | 0 |
| Conv2d_1 (Conv2D) | (None,37,37,64) | 18,496 |
| max_pooling2d_1 (Maxpooling2d) | (None, 18,18,64) | 0 |
| Conv2d_2 (Conv2d) | (None, 18,18,128) | 73,856 |
| max_pooling2d_2 (Maxpooling2d) | (None, 9,9,128) | 0 |
| Conv2d_3 (Conv2d) | (None, 9,9,128) | 147,584 |
| max_pooling2d_3 (Maxpooling2d) | (None,4,4,128) | 0 |
| Flatten (Flatten) | (None,2048) | 0 |
| Dense (Dense) | (None,512) | 1,049,088 |
| Dense_1 (Dense) | (None,2) | 1,026 |

## 3. Materials and Methods

The field surveys were conducted within the Pandeiros Environmental Protection Area (APA Rio Pandeiros), situated in the municipality of Bonito de Minas, northern Minas Gerais, Brazil (50º 85' 95, 36" O) (83º 16' 64, 18"). Target sites were delimited based on the historical incidence of pequi tree mortality and characteristic symptoms of shoot dieback, with field identification assisted by members of the local agroextractivist communities (Doerl, 2021).

Experimental monitoring was established across an area of approximately 100 hectares within the Água Doce community. Longitudinal assessments were carried out over a two-year period spanning four distinct sampling campaigns: January 2018 (rainy season), September 2018 (dry season), February 2019 (rainy season), and October 2019 (dry season). A total of 199 individual pequi trees were randomly selected and georeferenced (Figure 1). For each specimen, biometric attributes (such as trunk diameter) and phytosanitary metrics were recorded, including foliar spot incidence, stem borer infestation, and the incidence and severity of shoot dieback symptoms, as well as tree mortality rates (Doerl, 2021). Aerial imagery was systematically acquired across these plots using an Unmanned Aerial Vehicle (UAV).

In August 2023, a complementary field campaign was performed in the same study plots to acquire high-resolution multi-angle imagery of the cataloged trees using both UAV platforms and handheld smartphone cameras. The spatial distribution of the monitored specimens within the APA Pandeiros is illustrated in Figure 4.

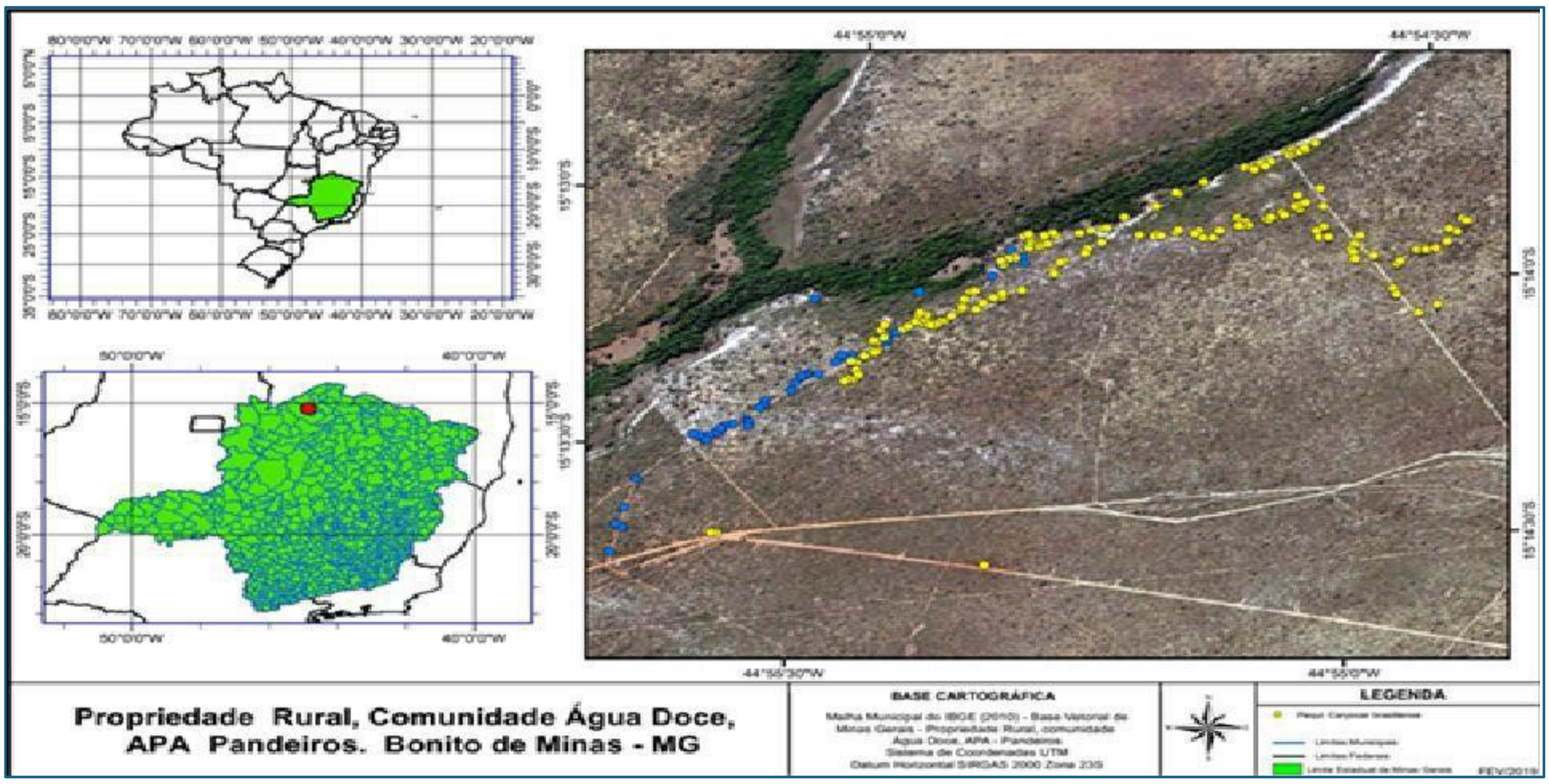

Fig.4 Yellow dots: pequi trees catalogued in the region. Source: (Doerl, 2021).

*3.1 Image acquisition*

To ensure dataset diversity and capture multi-scale pathological symptoms, imagery of pequi trees was acquired from varying perspectives, distances, and illumination conditions. Photographic records were standardized into three primary capture scales:

- Close-up view: High-resolution proximal captures focusing on localized foliar symptoms, structural bark lesions, and desiccated terminal twigs, enabling fine-grained extraction of textural and chromatic anomalies.

- Canopy (Full-tree) view: Intermediate perspective framing the entire crown structure, facilitating the assessment of branch architecture, overall defoliation patterns, and apical dieback.
- Landscape (Distance) view: Far-field captures contextualizing individual trees within the surrounding Cerrado canopy and soil matrix.

The collected raw dataset encompassed heterogeneous aspect ratios (1:1, 4:3, and 16:9) and native sensor resolutions, including $8{,}000 \times 6{,}000$, $4{,}928 \times 3{,}264$, $4{,}288 \times 2{,}848$, and $1{,}600 \times 758$ pixels, reflecting acquisitions from both the UAV platform and handheld mobile sensors. Representative captures across the three framing protocols are illustrated in Figure 5(a–c).

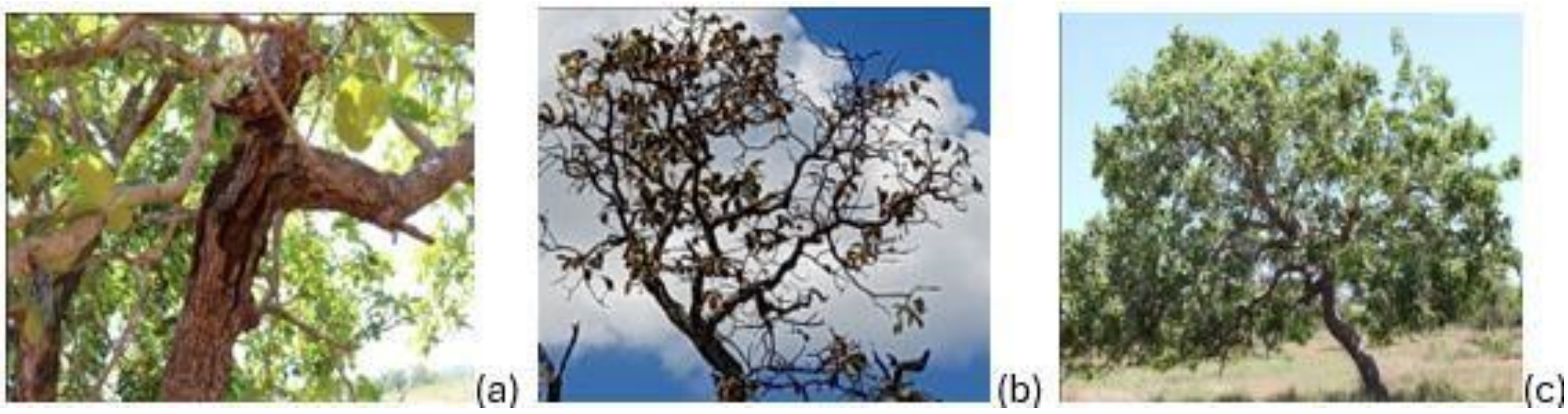

Fig.5 Examples of image capture conditions

*3.2 Data Augmentation Pipeline*

The curated raw dataset comprised 2,817 RGB images standardized to $512 \times 512 \times 3$ pixels. To mitigate class imbalance, enhance model generalization, and prevent overfitting during training, a synthetic data expansion pipeline was established, yielding a balanced corpus of 6,272 images. The augmentation strategy incorporated both geometric transformations and photometric adjustments applied to the training set. Specifically, transformations included random rotations ($\pm 15°$), horizontal and vertical flipping, spatial translation, scaling and zooming, and channel shifting (color jittering across RGB channels). These operations exposed the networks to diverse illumination profiles and varied canopy orientations. Representative visual outputs generated by each augmentation transformation are depicted in Figure 6.

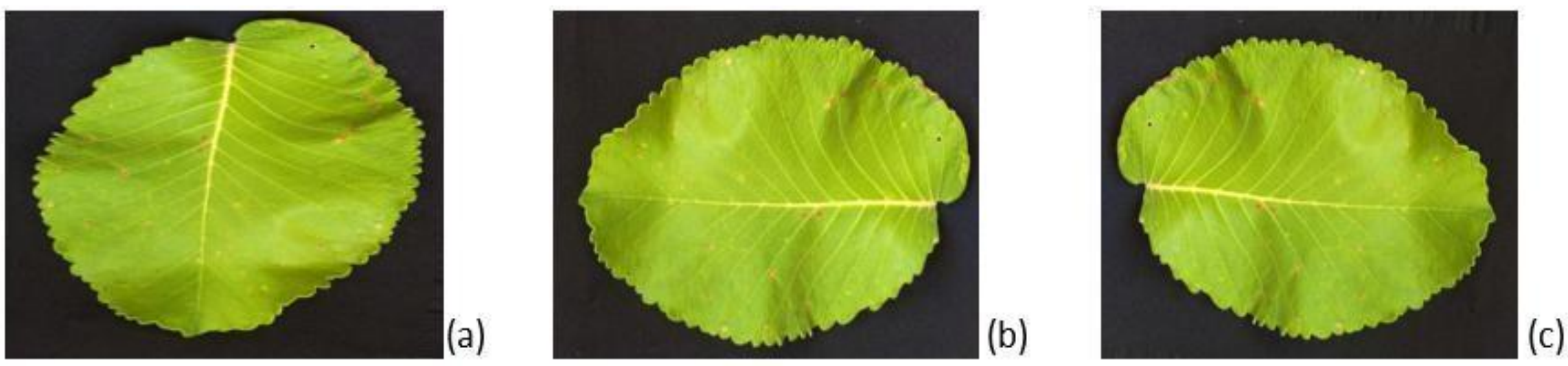

Figs. 6 (a), (b), and (c): Images of pequi leaves after different data enhancement processes
(a) original image, (b) image with rotation, and (c) image with inversion.

*3.3 Image Pre-processing and Normalization*

Following data augmentation, all images underwent standardization and intensity normalization prior to being fed into the neural networks. Pixel intensity values, originally ranging from 0 to 255 in the integer RGB space, were rescaled to the continuous interval $[0,1]$ (via division by 255.0). This rescaling standardizes the dynamic range across all input tensors, preventing gradient explosion or vanishing issues and accelerating backpropagation convergence during optimization. Furthermore, images were resized to conform to the native input dimensions required by each evaluated CNN architecture.

*3.4 Classification Workflow*

Developing an automated vision-based diagnostic tool provides an objective mechanism to identify pathological patterns associated with pequi tree mortality under Cerrado field conditions, ultimately supporting species conservation and targeted management strategies. The proposed end-to-end classification pipeline was structured into five interconnected stages:

- *Phase I*: Image Acquisition and Curation - Field imagery was collected across APA Rio Pandeiros using both UAV platforms and smartphone sensors under diverse illumination angles and backgrounds. The resulting dataset comprised both healthy specimens and trees exhibiting varying stages of shoot dieback. Metadata regarding sensor type, acquisition date, and environmental conditions were systematically logged.

- *Phase II*: Data Preprocessing and Partitioning - Raw captures were standardized in spatial dimensions and normalized across pixel channels ($[0,1]$). The dataset was then partitioned into independent training, validation, and testing subsets at the specimen level to prevent data leakage. Targeted data augmentation—including random rotations, flipping, and color adjustments—was strictly applied to the training set to address class imbalance.

- *Phase III*: Architectural Adaptation - Five benchmark convolutional neural network backbones (VGG16, LeNet-5, Inception, AlexNet, and MobileNet) were implemented. For each backbone, the feature extraction stages were retained while dense classification heads were customized for binary diagnosis (healthy vs. diseased).

- *Phase IV*: Model Training and Optimization - Networks were trained using categorical/binary cross-entropy loss and optimized via adaptive gradient algorithms (such as Adam). Hyperparameters—including batch size, initial learning rate, and learning rate scheduling—were tuned empirically using the validation split.

- *Phase V:* Evaluation and Diagnostic Robustness - Performance generalization was assessed on the isolated test set using standardized evaluation metrics: overall accuracy, precision, recall, F1-score, confusion matrices, and the Area Under the Receiver Operating Characteristic Curve (ROC-AUC).

*3.5 Evaluation metrics*

To quantitatively evaluate the classification performance of the evaluated CNN architectures, a confusion matrix was computed, and standard diagnostic metrics were derived based on VP (True Positive), VN (True Negative), FP (False Positive) and FN (False Negative). In this binary setup, infected plants (exhibiting shoot dieback) were defined as the positive class, whereas healthy plants constituted the negative class.

The accuracy measures the proportion of correctly classified instances relative to the total number of evaluated samples;

$$Accuracy = \frac{VP + VN}{VP + VN + FP + FN}$$

The sensitivity (recall) quantifies the model's ability to correctly identify diseased plants among all actual positive cases;

$$Sensitivity = \frac{VP}{VP + FN}$$

The specificity reflects the proportion of actual healthy plants correctly identified as negative;

$$Specificity = \frac{VN}{VN + FP}$$

The precision evaluates the proportion of true positive predictions among all instances predicted as diseased.

$$Precision = \frac{VP}{VP + FP}$$

The F1-Score is the harmonic mean of Precision and Sensitivity, providing a balanced diagnostic metric that accounts for class imbalances

$$F1-Score = 2 * \frac{Precision * Sensitivity}{Precision + Sensitivity}$$

Additionally, the ROC curve was plotted to evaluate diagnostic capability across varying decision thresholds, mapping the (true positive rate) and specificity (false positive rate) at different decision thresholds. True Positive Rate (TPR or Sensitivity):

$$TPR = \frac{TP}{TP + FN}$$

False Positive Rate (FPR):

$$FPR = \frac{FP}{FP + TN}$$

## 4. Analysis and Experimental Results

The experimental results were obtained using the image classification approach. Table 6 shows the comparative results of the models with any metrics. Table 7 and Fig 7 show the confusion matrices generated by the models with their respective analyses, followed by the ROC Curve for each model, generated by the results with their respective analyses.

### *4.1 Implementation of CNNs*

All CNN models were implemented in Python 3.11 using the TensorFlow 2.x and Keras frameworks, supported by Scikit-learn for metric computation and validation pipelines. Experiments were conducted within the Google Collaboratory environment provisioned with an NVIDIA A100 GPU (40 GB VRAM) running under an enterprise Linux environment.

### *4.2 Comparative Model Performance*

The comprehensive diagnostic performance across the five evaluated architectures is summarized in Table 6, with corresponding confusion matrices and ROC curves depicted in Figure 7 and Figure 8, respectively.

- Overall Accuracy: AlexNet achieved the highest overall accuracy at 64.4%, followed closely by MobileNet (63.5%) and LeNet-5 (62.2%). In contrast, VGG16 recorded the lowest accuracy (50.5%), aligning with baseline random assignment for a balanced binary evaluation.

- Precision and Sensitivity (Recall): MobileNet exhibited the highest Precision (0.71), indicating superior reliability when diagnosing positive (diseased) cases. AlexNet and MobileNet provided the highest Sensitivity values (0.50 and 0.44, respectively). Conversely, VGG16 exhibited zero sensitivity and zero precision (0.0), failing to detect any true positive instances. This behavior indicates that VGG16 suffered from optimization stagnation, collapsing into a trivial decision boundary where all samples were classified into the majority/negative class.

- Specificity: Both VGG16 and AlexNet yielded a specificity of 1.0 (no false positives recorded), while MobileNet and Inception achieved 0.82. LeNet-5 recorded the lowest specificity at 0.77.

- Model Complexity and Parameter Footprint: A substantial disparity in computational footprint was observed. Inception exhibited the largest overhead, totaling 228,410,648 parameters (76,136,882 trainable). In contrast, the adapted LeNet-5 represented the most lightweight topology (183,080 total and 61,026 trainable parameters), while MobileNet demonstrated a balanced trade-off between compact parameter representation and high diagnostic precision.

. **Table 6:** Data Results

| **Models** | **Accuracy** | **Parameters** | **Trainable** | **Specificity** | **Precision** | **Sensitivity** |
|---|---|---|---|---|---|---|
| VGG16 | 0.505 | 131,959,664 | 43,986,554 | 1.0 | 0.0 | 0.0 |
| Incepton | 0.581 | 228,410,648 | 76,136,882 | 0.82 | 0.65 | 0.33 |
| LeNet | 0.622 | 183,080 | 61,026 | 0.77 | 0.667 | 0.468 |
| AlexNet | 0.644 | 187,071,344 | 62,357,114 | 1.0 | 0.322 | 0.5 |
| MobileNet | 0.635 | 3,871,112 | 1,290,370 | 0.82 | 0.71 | 0.44 |

Confusion matrix analysis and computational runtimes provide deeper insights into the diagnostic capabilities and operational constraints of each architecture (Table 7 and Figure 7). The total test set comprised 1,881 evaluated instances (951 healthy and 930 diseased specimens).

- VGG16: Exhibited an extreme pathological bias toward the negative class, correctly identifying all 951 healthy instances (TN = 951), but completely failing to detect any diseased specimens (TP = 0, FN = 930). This outcome confirms a complete model collapse into trivial majority prediction, rendering it unusable for early phytosanitary screening. Furthermore, it required the longest runtime (2 min 59 s / 179 s), demonstrating high parameter redundancy without diagnostic utility.
- Inception: Achieved a more balanced distribution, correctly classifying 786 healthy specimens (TN = 786) and 307 diseased instances (TP = 307). Nonetheless, the occurrence of 623 false negatives (FN = 623) indicates a substantial deficiency in recognizing early or subtle foliar necrosis. Its total execution runtime was 2 min 52 s (172 s).
- LeNet-5: Benefiting from its minimalist topology, LeNet-5 recorded the shortest runtime (2 min 48 s / 168 s). In terms of class discrimination, it identified 734 TN and 436 TP, achieving a comparatively well-balanced trade-off between classes. However, 494 false negatives and 217 false positives underscore residual limitations in distinguishing ambiguous background textures from necrotic lesions.
- AlexNet: Yielded classification metrics comparable to Inception, registering 794 TN and 336 TP. While it produced fewer false alarms (FP = 157) than LeNet-5, the high false-negative count (FN = 594) constrained its diagnostic sensitivity. Its execution runtime was 2 min 54 s (174 s).

- MobileNet: Demonstrated the most competitive performance, correctly identifying 782 healthy plants (TN = 782) and 414 diseased specimens (TP = 414). By reducing false negatives to 516 (outperforming Inception and AlexNet) while maintaining moderate false positives (FP = 169), MobileNet demonstrated superior feature extraction efficiency with an execution runtime of 2 min 55 s (175 s).

**Table 7**. Data Results

| Models | Confusion Matrix | Execution times |
|---|---|---|
| VGG16 | [[951 0] [930 0]] | 2:59 min |
| Incepton | [[786 165] [623 307]] | 2:52 min |
| LeNet | [[734 217] [494 436]] | 2:48min |
| AlexNet | [[794 157] [594 336]] | 2:54 min |
| MobileNet | [[782 169] [516 414]] | 2:55 min |

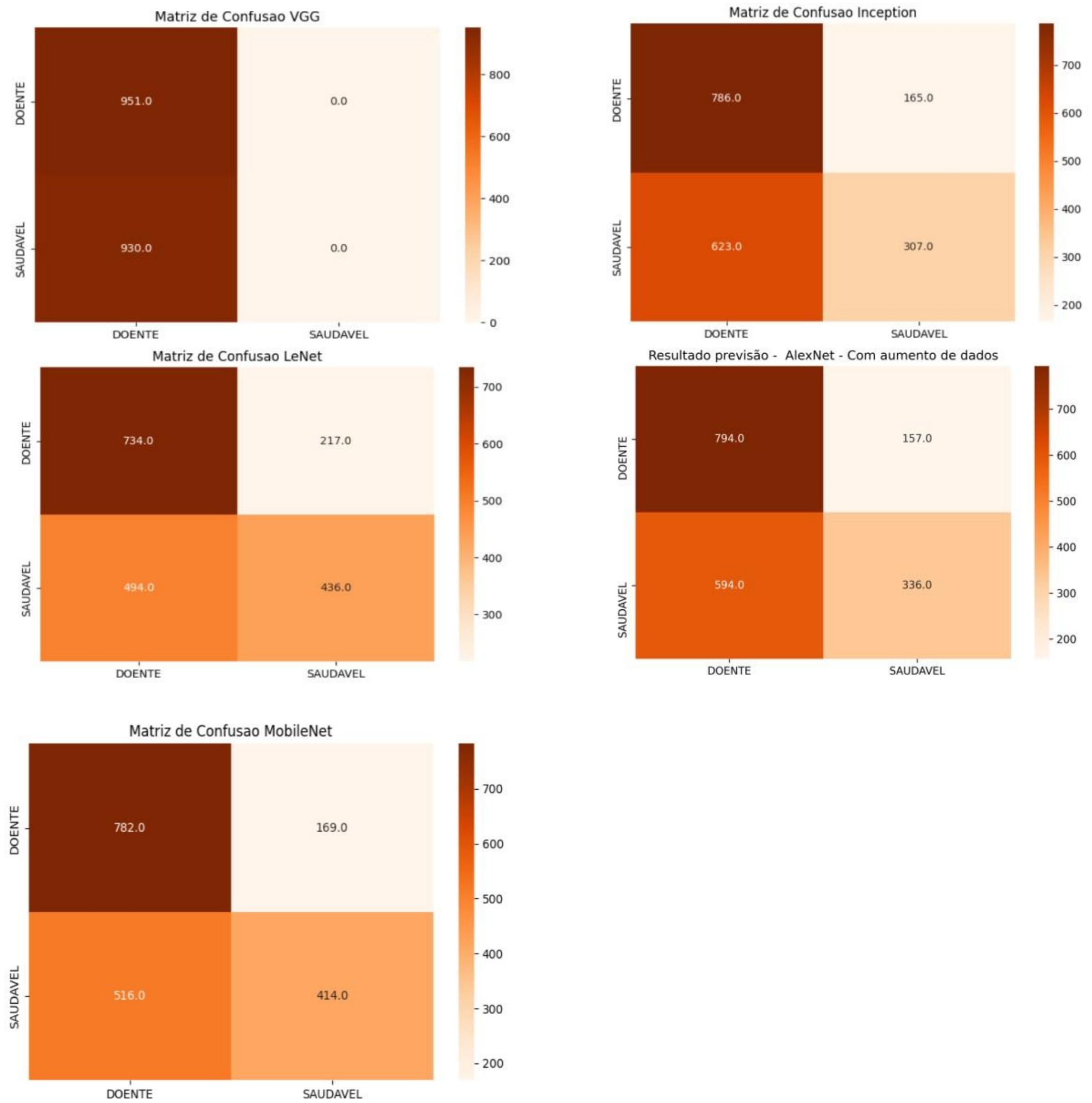

Fig 7. Results of the confusion matrices generated by the application according to the model

The ROC curves and corresponding AUC metrics illustrate the global discriminative capability across all decision thresholds for the five evaluated models (Figure 8):

- MobileNet (AUC = 0.63): Achieved the highest global discriminative performance among the evaluated architectures, exhibiting the most favorable trade-off between the True Positive Rate (TPR) and False Positive Rate (FPR) across operational thresholds.
- LeNet-5 (AUC = 0.62): Demonstrated the second-best performance, displaying competitive class separability despite its compact topology and minimal parameter footprint.
- AlexNet (AUC = 0.60): Maintained moderate discriminatory capacity, performing slightly below LeNet-5 across intermediate decision thresholds.
- Inception (AUC = 0.58): Exhibited marginal separability, performing only slightly above the random baseline (AUC = 0.50) and reflecting substantial false negative rates at typical operating points.
- VGG16 (AUC = 0.00): Produced a degenerate ROC profile resulting from complete classifier collapse. Because the network failed to predict any true positive instances (TP = 0), the model was entirely incapable of discriminating against diseased and healthy specimens.

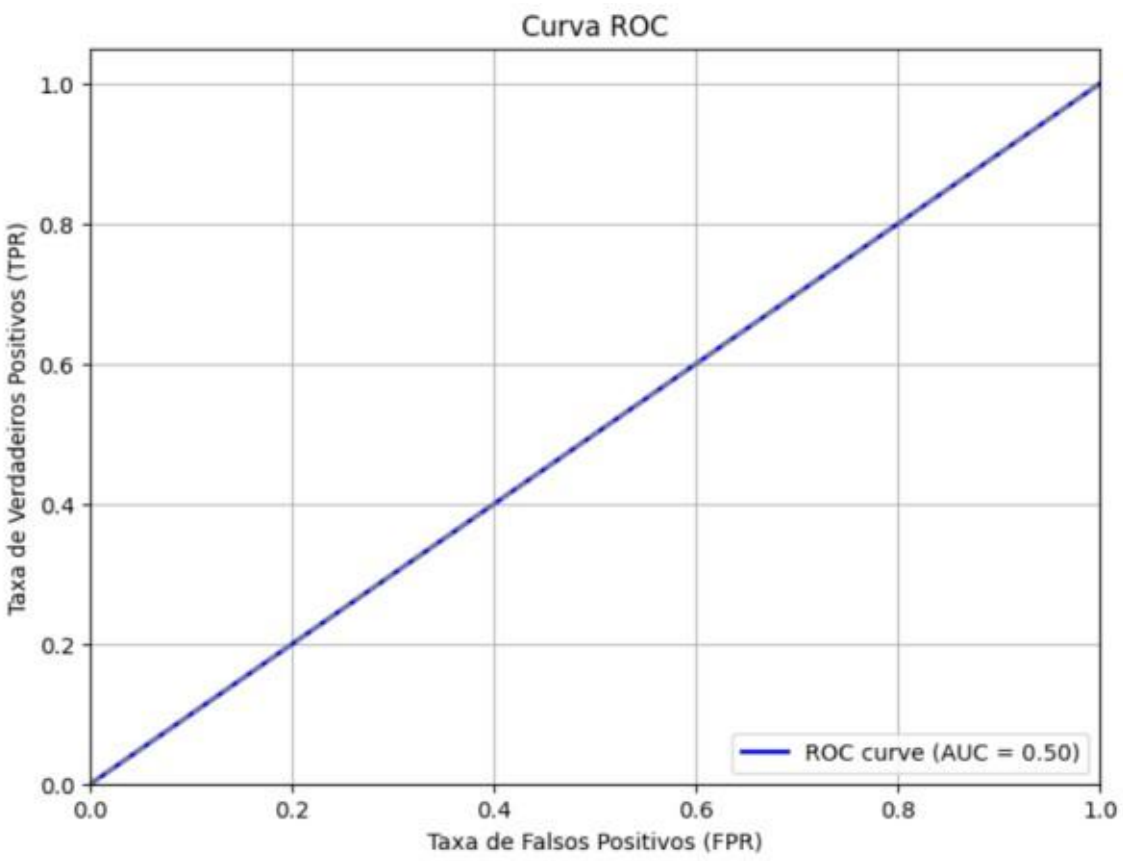

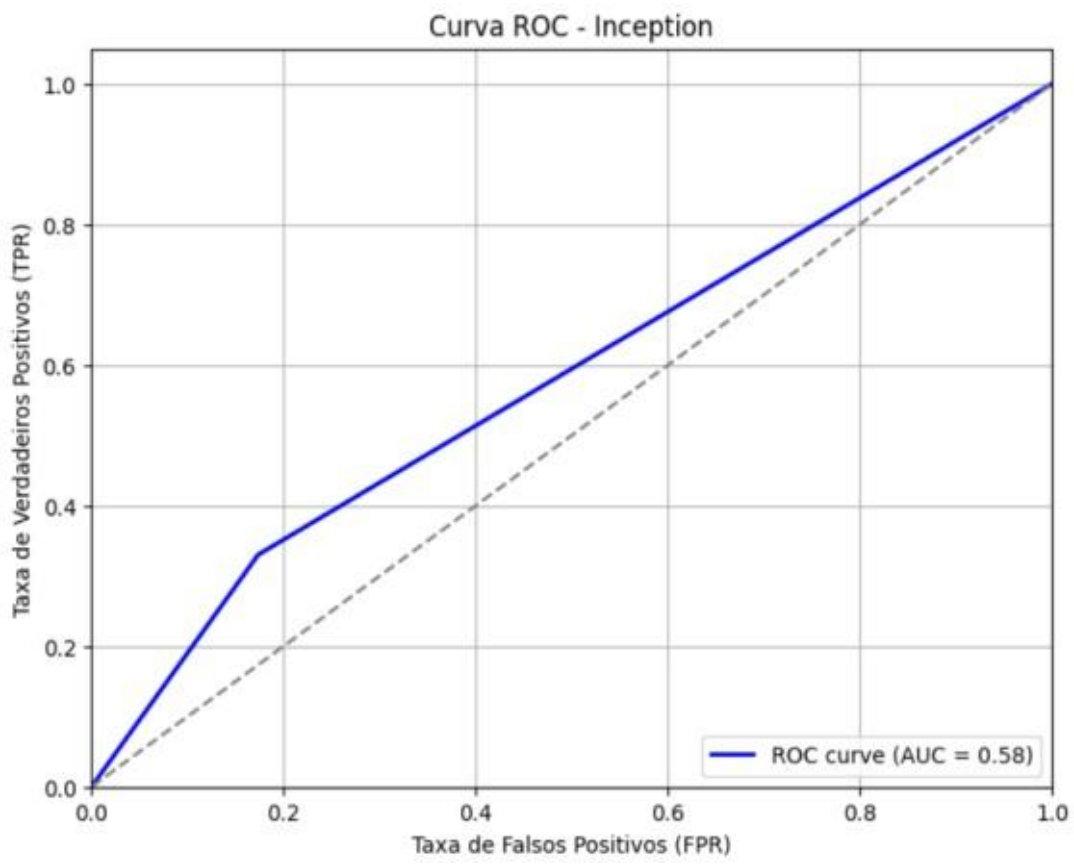

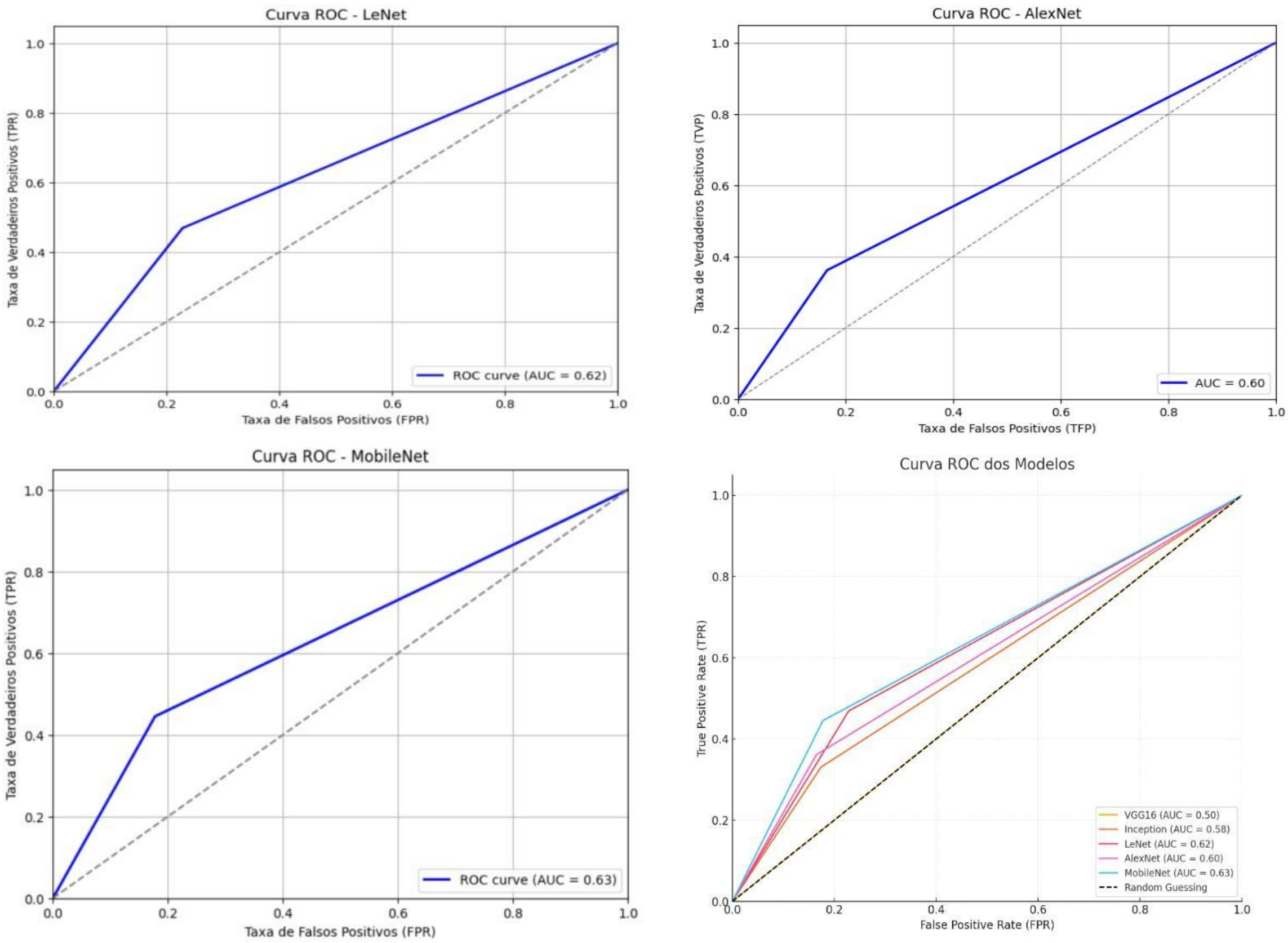

Fig 8. ROC curves showing the performance of five classification models (VGG16, Inception, LeNet, AlexNet and MobileNet).

*4.3 Discussion and Comparative Analysis*

The comparative evaluation reveals distinct operational trade-offs across the benchmarked architectures, highlighting the challenges of diagnosing shoot dieback under heterogeneous field conditions. Notably, architectural complexity did not correlate with improved diagnostic performance. Highly parameterized models, such as VGG16 and Inception, struggled with severe class imbalance and subtle symptomatic features. In particular, VGG16 collapsed entirely toward the majority negative class, yielding perfect specificity but complete failure in disease detection ($\text{Sensitivity} = 0.0$). Similarly, Inception exhibited low sensitivity relative to its substantial parameter footprint (over 228 million total parameters), indicating that excessive capacity without specialized transfer learning leads to severe optimization bottlenecks.

In contrast, lightweight topologies—specifically LeNet-5 and MobileNet, demonstrated superior operational viability for agricultural deployment. Benefiting from a compact parameter representation, LeNet-5 and MobileNet achieved a more balanced trade-off between sensitivity and specificity, while minimizing computational latency (recording runtimes under 2 min 55 s). MobileNet established the most favorable global profile, achieving the highest Precision (0.71) and ROC-AUC (0.63), followed closely by LeNet-5 ($\text{AUC} = 0.62$). These findings suggest that for field-level leaf and branch inspection, compact feature extractors that prioritize low- and mid-level texture representations are less prone to catastrophic overfitting than excessively deep backbones.

Nevertheless, all evaluated architectures exhibited elevated false negative rates, reflecting the intrinsic difficulty of identifying early-stage terminal necrosis and subtle defoliation patterns against complex Cerrado canopies. While synthetic data augmentation assisted in exposing the models to orientation and illumination variations, residual ambiguities between healthy aged foliage and initial necrotic lesions persisted. Addressing these diagnostic limitations in practical decision-support systems will require targeted methodological refinements, including attention mechanisms to localize necrotic foci, fine-grained multi-scale patch extraction, and the incorporation of multi-spectral or temporal field imagery.

## 5. Conclusion and Future Work

As part of an ongoing research initiative, this study investigated the feasibility of deep convolutional neural networks for automated visual diagnosis of branch dieback in pequi trees (*Caryocar brasiliense*), a severe phytopathology threatening both the conservation and agroextractivist sustainability of the Brazilian Cerrado biome. By benchmarking five distinct topologies, AlexNet, Inception, LeNet-5, MobileNet, and VGG16, against field imagery acquired under uncontrolled conditions, this work demonstrates the operational trade-offs between architectural complexity and diagnostic reliability.

The empirical findings reveal that heavily parameterized networks do not necessarily translate to superior performance in complex agricultural field settings. Notably, VGG16 suffered complete classifier collapse toward the majority healthy class, and Inception exhibited low sensitivity relative to its substantial computational overhead. Conversely, lightweight models demonstrated greater stability: MobileNet achieved the most favorable diagnostic profile, yielding the highest Precision (0.71) and ROC-AUC (0.63) closely followed by the compact LeNet-5 (AUC = 0.62, runtime of 2 min 48 s). These results indicate that compact convolutional feature extractors are more resilient against overfitting when distinguishing subtle foliar and terminal shoot necrosis under natural illumination.

This study highlights key findings and operational limitations that characterize the baseline stage of this research. We expect that these initial results will inspire future investigations to explore and integrate emerging technologies such as deep neural networks, explainable AI, embedded systems, and curated soil databases (Marinho et al., 2023). As this project progresses, future work will focus on the deployment of these technologies across mobile platforms, robotic systems, and UAVs, expanding the reach of intelligent monitoring for native species in remote and hard-to-access areas. This combination of technical innovation and rigorous agronomic validation can contribute significantly to sustainable extractivist practices, fostering the long-term conservation and preservation of this vital Brazilian biome.

## 6. Acknowledgments

This work was supported by the Coordination for the Improvement of Higher Education Personnel – Brazil (CAPES) – Funding Code [001] No. 88887.132576/2025-00; it also received partial support from the National Council for Scientific and Technological Development (CNPq) – Grant Nos. [310452/2025-2, 150731/2024-8, and 131524/2025-9] and FAPERJ – Grant No. [E-26/260.253/2026] and Alda Research Institute.